\pdfoutput=1

\documentclass[11pt]{article}

\usepackage[]{naacl2021}

\usepackage{times}
\usepackage{latexsym}

\usepackage[T1]{fontenc}

\usepackage[utf8]{inputenc}

\usepackage{microtype}

\usepackage{graphicx}

\usepackage{color, colortbl}
\usepackage{booktabs} 
\usepackage{multirow}
\usepackage{wrapfig}
\usepackage{appendix}
\usepackage{amsmath}
\usepackage{hyperref}
\usepackage{url}

\usepackage[linesnumbered,ruled,vlined]{algorithm2e}
\usepackage[utf8]{inputenc} 
\usepackage[T1]{fontenc}    
\usepackage{url}            
\usepackage{booktabs}       
\usepackage{amsfonts}       
\usepackage{nicefrac}       
\usepackage{microtype}      
\usepackage{xcolor}         
\usepackage{float}
\usepackage{graphicx}
\usepackage{hyperref}       
\usepackage{caption}
\usepackage{lipsum}  
\usepackage{float}
\usepackage{fancyvrb}
\usepackage{listings}
\usepackage{framed}
\title{NeuGPT: Unified multi-modal Neural GPT}

\author{\parbox{16cm}
  {\centering
    {Yiqian Yang$^{1}$\thanks{These authors have contributed equally to this work. Yiqian Yang initiated the project, designed the architecture, wrote the main code, trained and tested NeuTokenizer, and drafted the core content of the paper. Yiqun Duan significantly helped formalize the project, trained and tested NeuGPT, and was instrumental in authoring a substantial portion of the manuscript. }  \ \ \ \ \ \ \ Yiqun Duan $^{2*}$\ \ \ \ \ \ \  Hyejeong Jo $^{3}$ \ \ \ \ \ \ \ Qiang Zhang$^{1}$  \ \ \ \ \ \ \ Renjing Xu$^{1\dagger}$\ \ \ \ \ \ Oiwi Parker Jones$^{4\dagger}$ \ \ \ \ \ \ \ Xuming Hu$^{2\dagger}$ \ \ \ \ \ \ \ Chin-teng Lin$^{2\dagger}$ \ \ \ \ \ \ \ Hui Xiong$^{1}$\thanks{These are corresponding authors.\\
    1 The Hong Kong University of Science and Technology (Guangzhou), People's Republic of China\\
    2 GrapheneX-UTS HAI Centre, Australia Artificial Intelligence Institute, University of Technology Sydney, Australia\\
    3 Department of Software Convergence, Kyung Hee University, Republic of Korea\\
    4 PNPL, Department of Engineering Science, University of Oxford, United Kingdom\\
    Yiqian Yang: yyang937@connect.hkust-gz.edu.cn,
    Yiqun Duan: duanyiquncc@gmail.com,
    Hyejeong Jo: girlsending0@khu.ac.kr,
    Qiang Zhang: qzhang749@connect.hkust-gz.edu.cn,
    Renjing Xu: renjingxu@hkust-gz.edu.cn,
    Oiwi Parker Jones: oiwi@robots.ox.ac.uk,
    Xuming Hu: xuminghu@hkust-gz.edu.cn,
    Chin-teng Lin: chin-teng.lin@uts.edu.au,
    Hui Xiong: xionghui@hkust-gz.edu.cn
    } 
    }\\ 
  }
}

\begin{document}
\maketitle

\begin{abstract}

This paper introduces NeuGPT, a groundbreaking multi-modal language generation model designed to harmonize the fragmented landscape of neural recording research. Traditionally, studies in the field have been compartmentalized by signal type, with EEG, MEG, ECoG, SEEG, fMRI, and fNIRS data being analyzed in isolation. Recognizing the untapped potential for cross-pollination and the adaptability of neural signals across varying experimental conditions, we set out to develop a unified model capable of interfacing with multiple modalities. Drawing inspiration from the success of pre-trained large models in NLP, computer vision, and speech processing, NeuGPT is architected to process a diverse array of neural recordings and interact with speech and text data. Our model mainly focus on brain-to-text decoding, improving SOTA from 6.94 to 12.92 on BLEU-1 and 6.93 to 13.06 on ROUGE-1F. It can also simulate brain signals, thereby serving as a novel neural interface. Code is available at \href{https://github.com/NeuSpeech/NeuGPT}{NeuSpeech/NeuGPT (https://github.com/NeuSpeech/NeuGPT) .} 

\end{abstract}

\newpage

\section{Introduction}
The brain stands as the cornerstone of human intelligence, enabling us to perceive and interact with the world around us. Despite extensive efforts to unravel its complexities, our understanding of the brain remains limited. While we have at our disposal a range of brain recording devices, the development of large-scale methodologies has been hindered by considerable challenges. Our objective is to pioneer the creation of a comprehensive model, underpinned by extensive datasets, to interpret and manage a spectrum of brain signals. This endeavor promises to deepen our comprehension of the brain and to harness the power of neuroprostheses, thereby enhancing the quality of life for individuals with severe impairments.


The boundaries between mind and machine are increasingly blurred as remarkable progress in brain-to-language conversion sets the stage for a new era in AI-neuroscience. Through advancements in both invasive and non-invasive neural recording technologies, our vision is to establish a unified framework capable of encoding and decoding various forms of brain recordings and tasks - a breakthrough poised to transform brain-computer interfaces.

\subsection{Invasive Neural Recording Approaches}

Invasive techniques, particularly those employing ECoG (electrocorticography), have yielded promising outcomes in the decipherment of brain signals. A progression of ECoG-based decoding papers \cite{herff2015brain, moses2016, moses2019, makin2020machine, anumanchipalli_2019_speech_synthesis_neural_spoken_sentences_ecog_bilstm_brain2speech} led to the first, pioneering demonstration of brain-to-text decoding in a paralyzed individual \cite{moses2021neuroprosthesis}. This groundbreaking work combined speech detection, word classification, and an external language model but was restricted to sentences constructed from a limited vocabulary of 50 words. 


In 2021, \citet{Willett_2021_ecog_brain2text_handwriting_rnn_lm_brain2text} achieved a surprising breakthrough by developing a system capable of decoding handwritten characters from microelectrode array recordings, utilizing a Recurrent Neural Network (RNN) in tandem with a language model. Building on this foundation, \citet{Metzger_2022_ecog_rnn_beam_search_gpt_brain2text} established a comprehensive pipeline in 2022 that translated brain signals directly into text, with the assistance of the GPT-2 language model.

The field has also seen recent progress in the realm of large-vocabulary decoding. \citet{Metzger_2023_ecog_hubert_birnn_brain2speech_brain2text_avatar} introduced a real-time decoding system capable of interpreting speech, text, sentiment, and facial expressions. Similarly, \citet{Willett_2023_ecog_speech_neuroprosthesis_rnn_brain2speech2text} and \citet{card2024} contributed to the direct translation of neural activity into text, adapting a prior handwriting model \cite{Willett_2021_ecog_brain2text_handwriting_rnn_lm_brain2text}. In a parallel development, \citet{Liu_2023_ecog_chinese_cnn_lstm_brain2speech_brain2pinyin} extended the scope of this research to logo-syllabic languages by devising a tripartite model for the decoding of Chinese language. Concurrently, \citet{feng2023high_seeg_chinese_brain2language} achieved text interpretation using SEEG recordings, further expanding the horizon of brain-to-language conversion.

\subsection{Non-invasive Neural Recording Approaches}

Non-invasive methods, while offering a less obtrusive approach to brain signal capture, pose distinct challenges in the realm of signal interpretation. The brain-to-speech system developed by Meta \cite{D_fossez_2023_meg_eeg_clip_pretrain_meta_brain2speech} has harnessed the power of contrastive learning with MEG and EEG data, showcasing its ability to match fixed-length segments of data across audio and brain modalities. \citet{ghazaryan2023trials_MEG_decode_written_text} have delved into the decoding of written text using MEG, albeit with a restricted vocabulary.

In the sphere of EEG-to-text conversion, \citet{wang2022open_aaai_eeg2text} have crafted a system capable of translating EEG signals at the word level into textual output, leveraging a pre-trained BART model \cite{lewis2019bart}. This seminal research has paved the way for follow-up studies, such as the work by \citet{duan2023dewave_brain2text} on DeWave, which integrates wav2vec \cite{baevski2020wav2vec_wav2vec2_origin} and a discrete codex for enhanced representation before employing BART for text generation. \citet{jo2024are} investigated the limitations of EEG-to-text models and concluded that all previous EEG-to-text models were ineffective. Since then, however, NeuSpeech \cite{yang2024decode_neuspeech} has utilized Whisper, a foundation model for Automatic Speech Recognition \cite{radford2023robust_whisper_model_originalpaper}, to decode MEG signals to text in an open-vocabulary context, achieving state-of-the-art results for MEG. MAD \cite{yang2024mad} has further advanced the field by improving decoding performance on previously unseen texts.

This summary highlights the swift progression of brain-to-language technologies. While invasive techniques currently lead in open-vocabulary interpretation capabilities, non-invasive methods are steadily advancing, and may soon offer more accessible and less invasive solutions for brain-computer interface applications.


\subsection{Multi-modal LLM}

The advent of ChatGPT \cite{openai2022chatgpt} marked a significant milestone in the development of Large Language Models (LLMs). ChatGPT demonstrated the capability of LLMs to engage in natural language conversations, leading to a surge in interest for integrating additional modalities into these models. The LLAVA\cite{liu2023llava} series pushed the boundaries by incorporating visual modalities into LLMs through embedding methods, allowing the models to process and generate text with visual context. However, it cannot generate visual content.

Building upon this foundation, SpeechTokenizer \cite{zhang2024speechtokenizer} and SpeechGPT \cite{zhang2023speechgpt} revolutionized the integration of speech modalities by transforming audio into a code format that could be seamlessly integrated into LLMs. This approach not only facilitated the processing of speech but also laid the groundwork for a more generalized approach to multi-modal inputs. AnyGPT \cite{zhan2024anygpt} extended the methodology of SpeechGPT \cite{zhang2023speechgpt} by encoding visual and audio modalities as code, thus making it increasingly straightforward to generate and input multi-modal content. Essentially, these models treat other modalities as tokens, indistinguishable from text, thereby greatly accelerating the development of multi-modal LLMs.

In the realm of neuroscientific research, recent endeavors have also begun to harness the power of LLMs for multi-modal interactions, enhancing the analytical capabilities of neuroimaging data. For instance, the study by \citet{shen2024neuro_fmri_LLM_QA} has dealt with fMRI signals, enabling complex functionalities such as multi-turn dialogue, complex reasoning, visual reconstruction, and conceptual localization within brain signals. Traditionally, each of these tasks would require a bespoke model, but with LLMs, a single model can address a multitude of tasks, underscoring the utility and versatility of these LLMs.

Moreover, there has been promising exploration into developing discrete tokenizers for large-scale pre-trained EEG data. Works like LaBram~\cite{jiang2024large_labram} have shown that it is feasible to train models capable of converting continuous signals into discrete tokens and back, with significant performance improvements observed in downstream classification tasks. This success suggests that the principles of discrete token modeling may be effective to be applied to LLMs.

In light of these advancements, we are inspired to leverage the experience gained from discrete token modeling and apply it to LLMs. By doing so, we aim to further enhance the capabilities of LLMs in processing and generating multi-modal content, thereby paving the way for even more sophisticated applications in the AI-neuroscience field.

\subsection{NeuGPT}

Our work, NeuGPT, is motivated by several key observations within the neural recording field. Firstly, the domain has been characterized by a fragmented approach, where each signal type—EEG, MEG, ECoG, SEEG, fMRI, fNIRS—has been investigated in isolation, with limited cross-pollination due to the perceived uniqueness of each signal type. Secondly, the flexibility of neural signals is vast; experimental setups vary widely, sensor coordinates are not universally transferable, individual head shapes and sizes differ, and sampling rates span a broad range. Thirdly, the plethora of tasks within the neural domain, each requiring a custom model design and the inability to leverage other datasets or a unified pre-trained model, has significantly impeded progress. In contrast, fields like NLP, CV, and Speech have benefited from multiple pre-trained large models capable of tackling a variety of complex tasks~\cite{floridi2020gpt,openai2022chatgpt,radford2021learning_clip_origin,liu2023llava,radford2023robust_whisper_model_originalpaper,zhang2024speechtokenizer,zhan2024anygpt,zhang2023speechgpt}.

With these considerations in mind, our ambitious goal shown in Figure~\ref{fig:future} is to create a model that can intake all forms of neural recordings and interact with speech, text, and vision modalities. Furthermore, this model could serve as a new interface for tasks such as sleep detection or motor control decoding, and even simulate an individual’s brain signals. The works' success has laid the groundwork for a unified neural multimodal model. We aim to achieve this by designing a neural signal tokenizer and constructing a dataset of instructional prompts for LLMs.

Our approach is two-phased. The first phase involves the analysis of neural signals, where we train a neural tokenizer composed of an encoder, quantizer, and decoder to facilitate the conversion between neural signals and neural tokens. The second phase is the training of the LLM, during which we build a multi-modal instructional dataset. This decoupling ensures that primary signal modeling and complex relationship modeling are handled independently. To keep our research focused and manageable, in this iteration, we explore the conversion of MEG signals to text and speech using LLMs. Future work will delve into the other modalities.

Our contributions are as follows: 
\begin{enumerate}
    \item We are the first to construct an LLM capable of both ingesting and generating neural signals, whereas previous models could only process neural signals as input.
    \item We propose a discrete modeling approach for neural signals that demonstrates promising results.
    \item We have outperformed the previous state-of-the-art in the MEG-to-text task.
\end{enumerate}

\section{Method}


\begin{figure*}
    \centering
    \begin{minipage}[t]{0.7\linewidth}
        \centering
        \includegraphics[width=\linewidth]{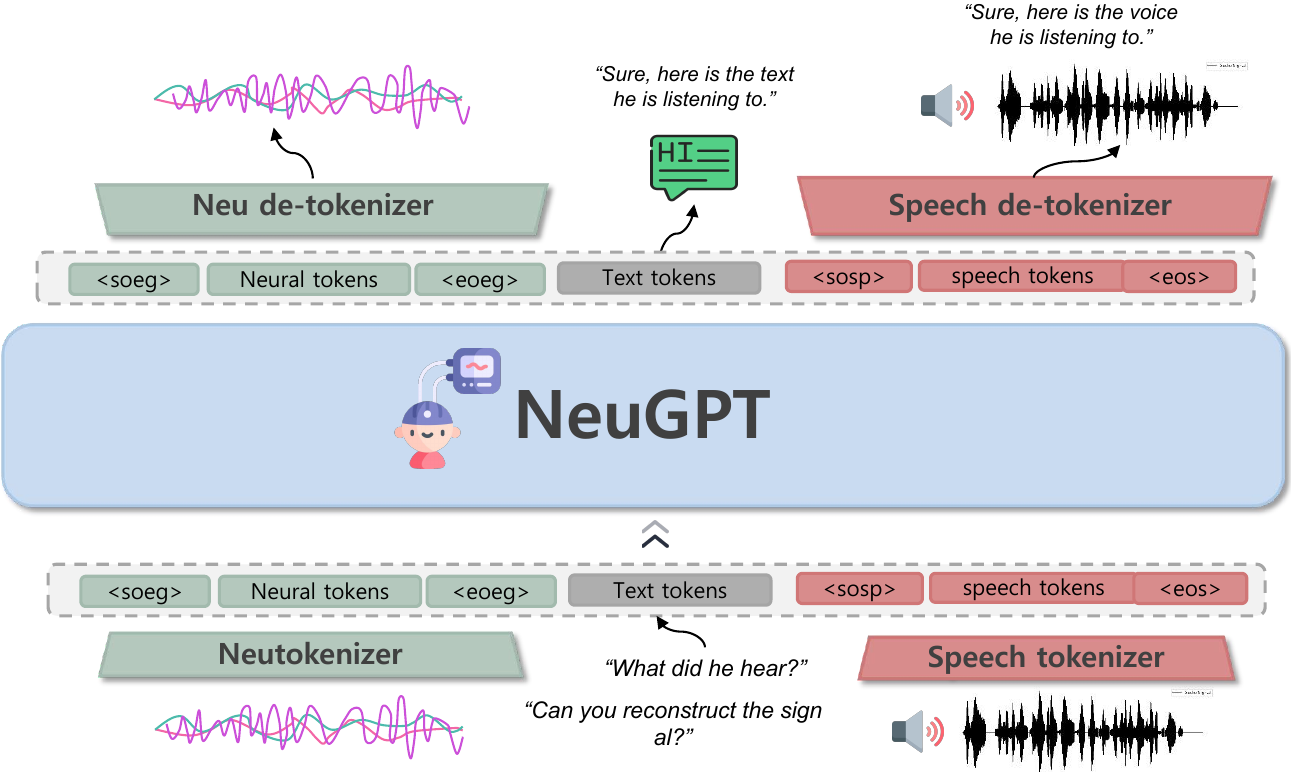}
    \end{minipage}
    \hfill
    \begin{minipage}[t]{0.28\linewidth}
        \centering
        \includegraphics[width=\linewidth]{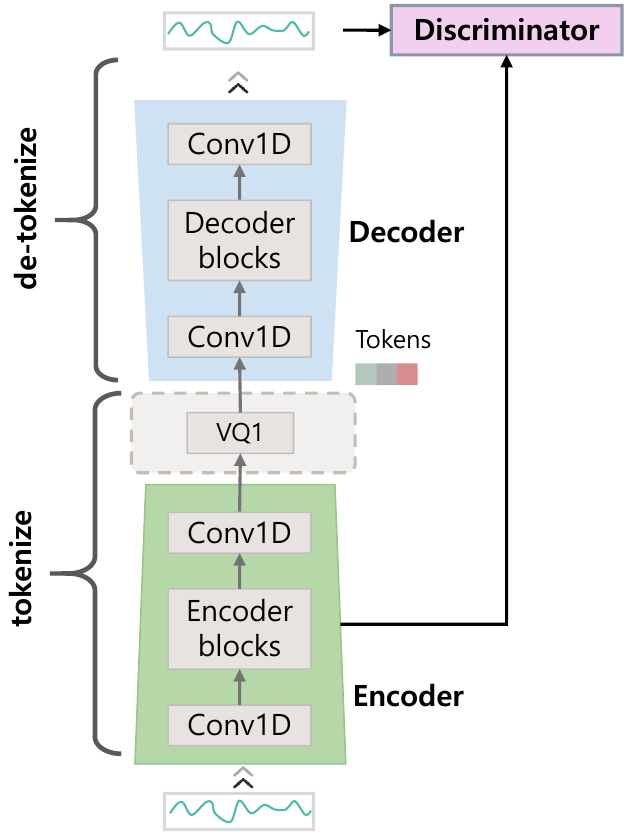}
    \end{minipage}
    \caption{(Left) NeuGPT overview: NeuGPT is designed to tackle the conversion between neural signals and other modalities such as text and speech. It uses 3 tokenizers for different modalities, converting all into tokens for the LLM to process. (Right) NeuTokenizer architecture: NeuTokenizer consists of an encoder, quantizer, decoder, and discriminator. The tokenizer is trained on a single channel, using the discriminator to improve synthetic results.}
    \label{fig:neugpt_overview}
\end{figure*}


Our research methodology leverages the power of Large Language Models (LLMs) to interpret and generate neural signals, inspired by recent advancements in multi-modal AI such as LLAVA \cite{liu2023llava}, SpeechGPT \cite{zhang2023speechgpt}, and AnyGPT \cite{zhan2024anygpt}. However, neural signals present unique challenges due to their complexity, variability, and sensitivity to recording conditions. To address these challenges, we propose a two-stage framework: (1) Neural Signal Tokenization, and (2) LLM Fine-tuning for Neural Code Understanding.

\subsection{Task Definition}

This study pursues performance increase in brain2text task. We have two main objectives for our study. The first is to create a robust neural tokenizer, referred to as $\Xi$ (NeuTokenizer), which is designed to effectively encode neural signals ($x$) into a discrete token format and decode these tokens back into neural signals. The second objective is to enhance the capabilities of a language learning model (LLM), denoted as $\Theta$, to generate content across various modalities with high fidelity.

For clarity, we define the following components for our task:
Neural signals are denoted by $x$.
Speech waveforms are represented by $s$.
Text is denoted by $t$.
The intermediate neural representation, post-encoding and pre-decoding, is $z$.
The neural code index after quantization is $c$.
The speech code index is $c_s$.
The text code index is $c_t$.

The operations performed by $\Xi$ are $x \overset{\Xi}{\rightarrow} c$ and $c \overset{\Xi}{\leftarrow} x$. Here, $x \overset{\Xi}{\rightarrow} c$ indicates the process of tokenizing neural signals into code indices, while $c \overset{\Xi}{\leftarrow} x$ indicates the reverse process of detokenizing code indices back into neural signals.

The generative function of the LLM $\Theta$ is represented by $\wp$. Our goal is to train $\Theta$ to excel in generating text with neural modality, enabling a seamless translation from neural data to linguistic output.


\subsection{Stage 1: Neural Signal Tokenization}

The first stage involves developing a neural tokenizer capable of encoding and decoding neural signals into discrete codes. We first considered LaBram~\cite{jiang2024large_labram} as a neural tokenizer, it can handle a relatively flexible layout, but it only handles standard 10-20 systems, and it is not good at generating neural signal directly, it predicts Fourier spectrum for each segment, so it is not continuous in the time dimension. Hence, we adopt the auto-encoder approach from \cite{zhang2024speechtokenizer}, customizing it for multi-channel neural signals. We tokenize neural signals on each single channel, hence it has great flexibility. Besides, it used both time domain, and spectrogram domain loss and used discriminator to make it more real and continuous.

\subsubsection{Architecture}

Our neural tokenizer is inspired by speechtokenizer~\cite{zhang2024speechtokenizer}, which is consists of four key components (Figure \ref{fig:neugpt_overview}):

1. Encoder: Transforms raw neural signals into embeddings.
2. Quantizer: Converts embeddings into discrete code indices.
3. Decoder: Reconstructs neural signals from the quantized embeddings.
4. Discriminator: Enhances the quality of reconstructed signals.


The Encoder and Decoder are from SEANET~\cite{tagliasacchi2020seanet}, and the Quantizer is a Residual Vector Quantizer~\cite{zeghidour2021soundstream_RVQ}. Neural signal of shape $[B,C,T]$ is encoded by the encoder into neural representations of shape $[B,R,T_E]$. This is then quantized by the Quantizer into discrete tokens of shape $[B,R]$. The tokens can be de-quantized back to neural representations of shape $[B,R,T_E]$, And the Decoder generates the neural signal $[B,C,T]$, where $B$, $C$, $T$, and $R$ represent the batch size, channel number, time length, and representation dimension, respectively. The neural tokenizer is able to tokenize continuous neural signal and de-tokenize to recover it.

\subsubsection{Training Objectives}

In this section, we detail the loss functions employed in the training of our generative model, which are crucial for producing high-quality and realistic outputs. The loss functions below described in this section are all based on SpeechTokenizer~\cite{zhang2024speechtokenizer}. For the sake of conciseness, we only cite them here.

\paragraph{Reconstruction Loss ($L_t$)} The reconstruction loss ensures that the generator can accurately reconstruct the input data. It is defined as the L1 norm between the original input $x$ and its reconstructed version $\hat{x}$:

\begin{align}
L_t = \|x - \hat{x}\|_1
\end{align}

\paragraph{STFT Loss ($L_f$)} The STFT loss encourages the generator to produce features in the generated images that closely match those of real images. This is computed as the sum of L1 and L2 norms of the differences in features at multiple scales:

\begin{align}
L_f = \sum_{i\in\epsilon} \|\mathcal{S}_i(x) - \mathcal{S}_i(\hat{x})\|_1 +\|\mathcal{S}_i(x) - \mathcal{S}_i(\hat{x})\|_2 
\end{align}

where $\mathcal{S}_i$ represents the STFT function at scale $i$.

\paragraph{RVQ Commitment Loss ($L_w$)} The RVQ commitment loss ensures that the quantized vectors are close to their committed vectors:

\begin{align}
L_w = \sum_{i=1}^{N_q} \|z_i - z_{qi}\|_2^2
\end{align}

where $z_i$ is the vector before quantization, and $z_{qi}$ is the quantized vector.

The discriminator loss $L_D$ is a critical component in the training of generative adversarial networks (GANs). It ensures that the discriminator can effectively differentiate between real and fake samples. The loss function is defined as follows:

\begin{equation}
\begin{split}
L_D &= \frac{1}{K} \sum_{k=1}^{K} \left( \max \left( 0, 1 - D_k(x) \right) \right. \\
&\qquad \left. + \max \left( 0, 1 - D_k(\hat{x}) \right) \right)
\end{split}
\end{equation}

where:,$K$ is the total number of discriminators, which can be used in a multi-scale or multi-period setup to enhance the discriminative power of the network. $D_k$ is the $k$-th discriminator in the network. $x$ represents a real sample from the data distribution. $\hat{x}$ is a generated sample produced by the generator.

The terms $\max(0, 1 - D_k(x))$ and $\max(0, 1 + D_k(\hat{x}))$ are known as the hinge loss, which encourages the discriminator to output values that are at least 1 unit away from the decision boundary for real and generated samples, respectively. This formulation helps to stabilize the GAN training process and prevents the discriminator from becoming too confident, which can lead to vanishing gradients for the generator.

\paragraph{Gradient Penalty Loss ($L_g$)} The gradient penalty loss is applied to stabilize the training of the discriminator:

\begin{align}
L_g = \frac{1}{K} \sum_{k=1}^{K} \max(1 - D_k(\hat{x}), 0)
\end{align}

\paragraph{Feature Consistency Loss ($\mathcal{L}_{feat}$)} This loss enforces feature consistency between real and generated samples:

\begin{align}
L_{feat} = \frac{1}{KL} \sum_{k=1}^{K} \sum_{l=1}^{L} \frac{\|D^l_k(x) - D^l_k(\hat{x})\|_1}{\text{mean}(\|D^l_k(x)\|_1)}
\end{align}

where $D^l_k$ is the $l$-th feature map extracted by the discriminator for the $k$-th sample.

\paragraph{Generator Loss ($\mathcal{L}_G$)} The overall generator loss is a weighted sum of the individual losses:

\begin{align}
L_G = \lambda_t L_t + \lambda_f L_f + \lambda_g L_g + \lambda_{feat} L_{feat} + \lambda_w L_w
\end{align}

The weights $\lambda_t, \lambda_f, \lambda_g, \lambda_{feat},$ and $\lambda_w$ are hyper-parameters that control the relative importance of each loss component.





\subsection{Stage 2: LLM Fine-tuning for Neural Code Understanding}

Inspired by AnyGPT~\cite{zhan2024anygpt}, in the second stage, we fine-tuned an LLM to understand and generate neural codes, enabling cross-modal communication between neural signals, speech, and text. We used QWEN2-1.5B~\cite{qwen} as LLM base to train and evaluate, since it is relatively small, and efficient to train, and it has context length of 32K, and it has good performance on many metrics.

\subsubsection{Vocabulary extending}
Code indices of different modalities have different text marks to identify modality, here id is a var. For speech, code indice is <id>, start and end token for speech is <sosp>, <eosp>. Neural code indice is <EGid>, we also have start and end token for neural, <soeg> and <eoeg>. Since neural signal has many channels, we want LLM to have a better understanding of channels and time, we tie tokens of all channels at each time step together in the order of channels and insert a <nts> token before each time step. These special marks for multi-modal are added as tokens in the text tokenizer.

\subsubsection{Dataset Construction}

We create a multi-modal instruction-tuning dataset that includes various combinations of neural signal tokens, speech, and text. This dataset encompasses six modality pairs to ensure comprehensive cross-modal understanding and generation. 

\subsubsection{Pretraining Process}

We pretrained our model using cross modal instruction data released by SpeechGPT~\cite{zhang2023speechgpt}, we changed the original format to chatml format. An example prompt structure is below, $...$ means ellipsis:

\begin{lstlisting}[breaklines]
system: You are a helpful assistant named NeuGPT. You can understand and produce neural signals, and you can interact with speech and text modalities.
user: Can you speak the text using an exaggerated accent? This is input: as the snake squeezed him tighter and tighter,
assistant: <sosp><334>...<77><332><334><eosp>

\end{lstlisting}

\subsubsection{Fine-tuning Process}

We fine-tune the LLM using prompt-based learning. We constructed conversion pair between 3 modalities (speech, text and neural), resulting in 6 modality pairs. Here, eg means neural signal. An example of neural-to-text conversion prompt structure is below, we only set the corresponding text in the generated content:

\begin{lstlisting}[breaklines]
system: You are a helpful assistant named NeuGPT. You can understand and produce neural signals, and you can interact with speech and text modalities.
user: Convert the following eg input to text:
This is the input:<soeg><nts><EG5792><EG7851><EG7851><EG7851><EG7851><EG8128><EG7386><EG5857><EG7343><EG7598><EG7851><EG3241><EG3663>...<eoeg> 
assistant: incomplete page before him. His pen flickered

\end{lstlisting}

This approach allows the model to learn the relationship between neural codes and their corresponding textual or speech representations.



\section{Experiments}
\subsection{Dataset}
\label{sec.dataset}
The MEG-MASC dataset~\cite{Gwilliams_2023_dataset_meg_208sensors_27persons_56h} is a magnetoencephalography (MEG) dataset designed for assessing natural speech comprehension. It features MEG recordings from 27 participants proficient in English. These participants engaged in two separate sessions, each involving one hour of listening to four stories, which are ``cable spool fort",``easy money",``lw1",``the black willow". To get a fair evaluation, we adopt same previous settings in MAD~\cite{yang2024mad}. We test on ``cable spool fort", validate on ``lw1" and train on other stories. Details are in Table~\ref{dataset_split_details}. 

\begin{table*}[!h]
    \centering
     \caption{Details about the dataset splits, we ensured the three splits are totally separated. Unique sentences means the sentences that are different with other sentences, same meaning for unique words. There is no overlap sentence between train and test set. 371(46\%) means 371 words in test set is also in train set, accounting for 46 percentage. }
\resizebox{1.0\linewidth}{!}{
    \begin{tabular}{cccccll}
    \toprule
         Split&  Segments&  Unique sentences&  Words& Unique words &Overlap sentence &Overlap words\\\midrule
         train&  133966&  13266&  150497&  2776& -&-\\
         validation&  14896&  1387&  156027&  478& -&-\\
         test&  31115&  3151&  355654&  805& 0&371(46\%)\\ \bottomrule
    \end{tabular}
   }
    \label{dataset_split_details}
\end{table*}

For preprocessing, we used a band pass filter the MEG signal $\varepsilon$ between 0.1 Hz and 85 Hz,  then it is resampled from 1000 Hz to 400 Hz to reduce computing. MEG data is calibrated and presented in units of 200fT. We ensure that we separated training, evaluation, testing set totally since we used one story for testing, another story for evaluation, last two ones for training. We extract 4-second windows from the MEG-speech-text pairs, sliding every second and randomly shifting the window by ±0.5 seconds to generate samples. 

\subsection{Implement Details}
The learning rate was set to ${1e}^{-4}$, and the model was optimized using the Adam optimizer with betas=[0.9,0.99]. The batch size is 512. The dictionary size for the neural tokenizer was set to 8192, and the encoder output was configured to 32 dimensions. The encoder reduces the data resolution, condensing every 100 time steps into a single time step.  For the STFT loss, we utilize five scales, in which the largest scale is configured with a window length of 512 and a hop length of 128. For each subsequent scale, the window length and hop length are halved, effectively reducing the scale size. We set $\lambda_t=500, \lambda_f=9, \lambda_g=1, \lambda_{feat}=1, \lambda_w=10$. 

We trained NeuGPT on 8 NVIDIA A100 GPUs for 5 epochs, utilizing a batch size of 2. The training parameters included a gradient accumulation step of 8, a learning rate of 3e-4, a weight decay of 0.01, an Adam optimizer with beta2 set to 0.95, a warm up ratio of 0.01, and a cosine learning rate scheduler. This training takes 5 days.

For the speech tokenizer, we followed SpeechGPT~\cite{zhang2023speechgpt}, using Hidden-unit BERT (HuBERT)~\cite{hsu2021hubert_orig} to process speech to tokens and using multi-speaker unit HiFi-GAN~\cite{polyak2021speech_unit_vocoder}.

\subsection{Metrics}
We mainly assessed the quality of generated outputs on text. For text outputs, we employed BLEU (Bilingual Evaluation Understudy) to measure the accuracy of the generated text against a set of reference translations, ROUGE (Recall-Oriented Understudy for Gisting Evaluation) to evaluate the quality of text summarization, BertScore, CER (Character Error Rate) and Self-BLEU to determine the accuracy of brain-to-text transcription. Here all metrics are calculated using TorchMetrics.

\subsection{Results}
\subsubsection{NeuTokenizer Reconstruction}

In this section, we present the capabilities of NeuTokenizer in reconstructing neural signals, as evidenced by both temporal and spectral analyses.

\begin{figure}[ht] \centering \begin{minipage}{\linewidth} \centering \includegraphics[width=1\linewidth]{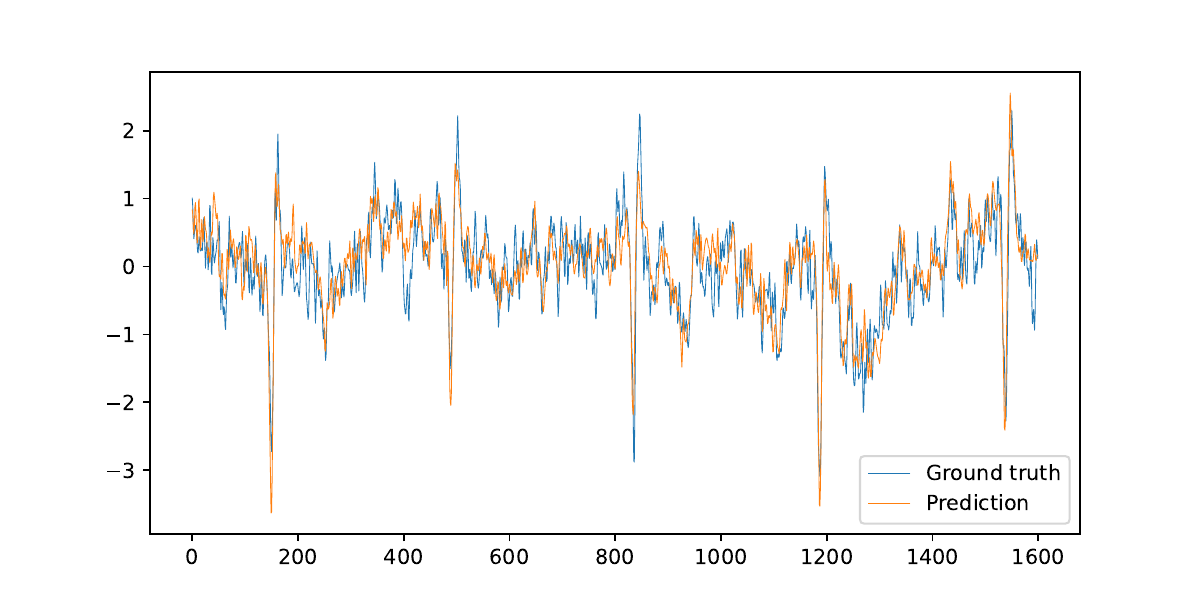} \caption{Temporal recovery of the neural signal by NeuTokenizer. The x-axis is the sample index.} \label{fig:neutokenizer_temporal_recovery} \end{minipage} \hfill 
\begin{minipage}{\linewidth} \centering \includegraphics[width=1\linewidth]{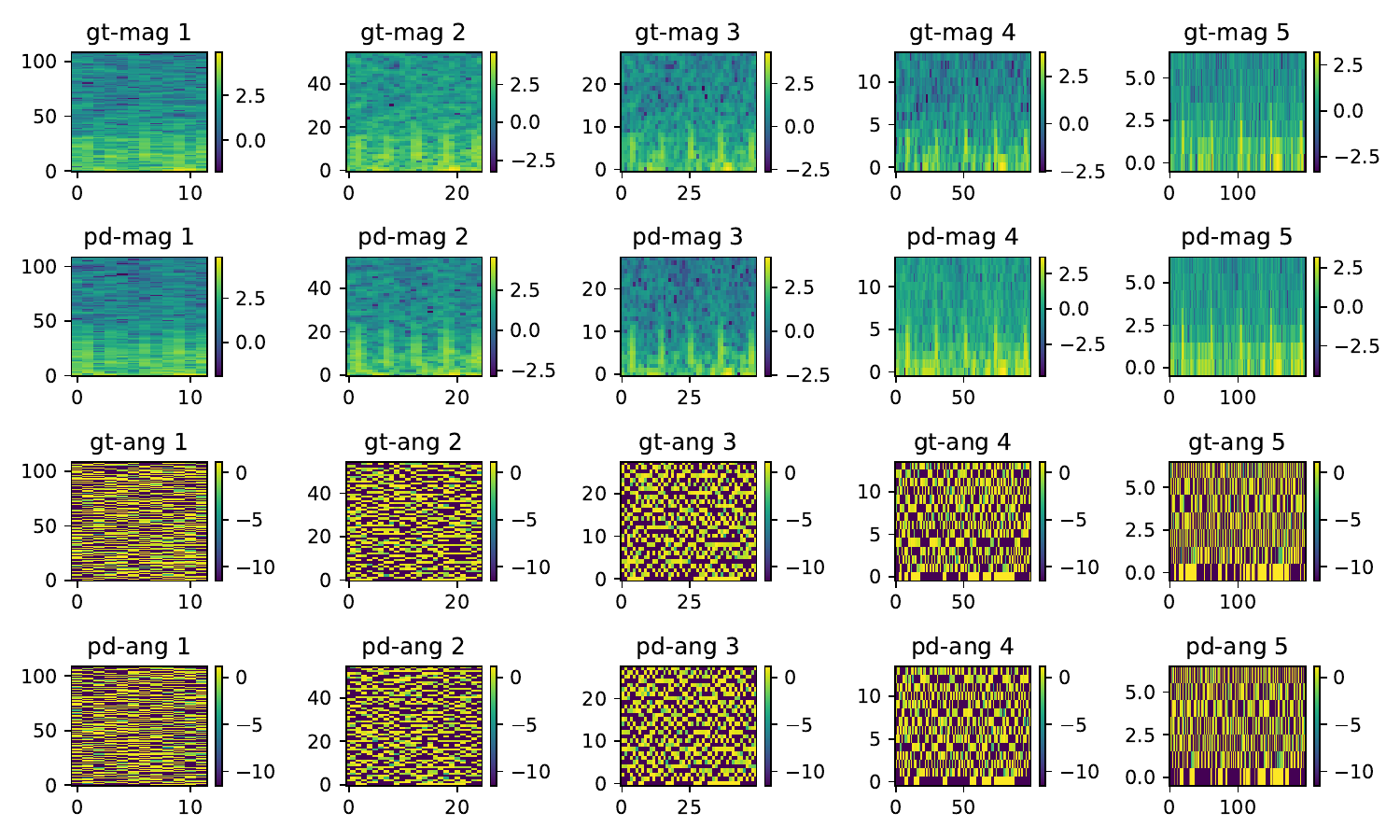} \caption{Spectral recovery of the neural signal by NeuTokenizer across different scales. gt means ground truth, pd means prediction, mag means magnitude, ang means angle. 1-5 means different scales of STFT.} \label{fig:neutokenizer_spectral_recovery} \end{minipage} \end{figure}

Figures \ref{fig:neutokenizer_temporal_recovery} and \ref{fig:neutokenizer_spectral_recovery} showcase the proficiency of our NeuTokenizer model in reconstructing neural signals. The temporal plot in Figure \ref{fig:neutokenizer_temporal_recovery} reveals a high-fidelity recovery of the signal in the time domain, indicating that NeuTokenizer maintains the integrity of the original neural activity patterns. Similarly, the spectral plot in Figure \ref{fig:neutokenizer_spectral_recovery} demonstrates the model’s efficacy across various frequency scales, with a clear and accurate representation of the signal’s spectral content. This dual-domain analysis confirms that NeuTokenizer is not only effective in preserving the temporal dynamics of neural signals but also in capturing the nuanced frequency characteristics, thereby offering a robust and comprehensive approach to signal recovery in neuroscientific applications.

\subsubsection{Text Generation}

\begin{table}[htbp]
\caption{Speech2Text performance of NeuGPT.}
\label{tab:speech2text_metrics}
\centering
\begin{tabular}{lcccccccc}
\hline
Metric &WER &BLEU-1 &ROUGE-1 Recall \\
\midrule
Value &10.35 &90.28 &91.12 \\
\hline
\end{tabular}
\end{table}

\begin{table*}[!h]
\caption{Brain-to-text conversion comparison. Random selecting is randomly selecting text from the test set.}
\label{compare_with_other_models}

\resizebox{1.0\linewidth}{!}{
\centering
\begin{tabular}{lccll}
\hline
Method           & BLEU-1(\%)$\uparrow$ & ROUGE-1F(\%)$\uparrow$ & BERTScore$\uparrow$ & CER$\downarrow$    \\ \hline
Random Selecting & 5.86                 & 7.20                    & 83.73     & 87.30  \\
NeuSpeech \cite{yang2024decode}         & 5.49                 & 8.43                    & 83.98     & 77.02  \\
Wav2vec2CTC~\cite{D_fossez_2023_meg_eeg_clip_pretrain_meta_brain2speech}      & 0.55                 & 1.44                    & 76.02     & 152.23 \\
MAD~\cite{yang2024mad}              & 6.94                & 6.93                    & 83.39     & 89.82  \\
NeuGPT           & 12.92                & 13.06                   & 83.62     & 99.8   \\ \hline
\end{tabular}

}
\end{table*}

\begin{table*}[!h]
    \centering
\caption{Brain2text generation. We show some generated sentences with the ground truth. \textbf{Bold} means exact match, and \textit{Italy} means fuzzy match.}
\label{tab:decoding text demos}
\begin{tabular}{@{}ll@{}}
\toprule
\multicolumn{2}{l}{Decoding results on MEG-MASC data~\cite{Gwilliams_2023_dataset_meg_208sensors_27persons_56h}}                                                       \\ \midrule
Prediction   & the ground. His hand blurred and snapped out as he \textbf{stood} back \textit{slilling}                               \\ \hline
Ground truth & \textbf{stood} stuck, \textit{still}, waiting for the first tiny gleam from the scout craft to appear in \\ \midrule
Prediction   & \textbf{the other} boys began to \textit{crowd around} at a respectful distance. Do you think he's dead?   \\ \hline
Ground truth & not to seem nervous. The sky met the flat ground \textit{in all directions} on \textbf{the other}          \\ \bottomrule
\end{tabular}

\end{table*}

We demonstrated the speech-to-text capabilities of NeuGPT to illustrate the practical utility of our model in straightforward tasks. It is the inherent challenges within non-invasive signals that constrain the high performance levels typically achievable in speech-to-text applications.

In the realm of brain2text generation, our model exhibits superior performance when contrasted with prior methodologies. The metrics we employed reveal a marked improvement in the quality and coherence of the generated text, surpassing the benchmarks set by all previous models. Our NeuGPT has a significant increase in ROUGE-1F score of 13.06\% compared to previous model MAD~\cite{yang2024mad}. We also demonstrated some predictions in Table.~\ref{tab:decoding text demos}. It has a relatively long set of correct words in the first example, and it has ``crowd around'' which is similar to ``in all directions'' in meaning. The precision and fluency of the text produced by our system underscore its advanced capabilities in capturing linguistic nuances and maintaining semantic consistency.





\section{Conclusion}

In conclusion, this study has traversed the multifaceted landscape of multi-modal language generation, leveraging the power of neural signals to bridge the gap between brain activity and expressive communication. Our introduction laid the groundwork for a novel approach to understanding and harnessing the intricate relationship between neural dynamics and linguistic expression. Through the development and application of our model, NeuGPT, we have not only demonstrated the technical feasibility of translating neural signals into coherent speech and text but also illuminated the path for more sophisticated interactions between the human brain and artificial intelligence.

The insights gained from this study are twofold. Firstly, we have shown that a unified framework for neural signal processing can indeed overcome the historical fragmentation within the field, where different signal types were often studied in isolation. Secondly, the flexibility of our approach, which accommodates the diverse layouts and sensor coordinates of neural recording, paves the way for a more generalized and adaptable model that can be applied across various experimental settings.

As we reflect on these findings, it becomes apparent that the integration of neural signals into language generation models not only advances the state-of-the-art in AI but also offers a deeper insight into the human brain’s language processing mechanisms. This work serves as a catalyst for future research, inspiring the exploration of more complex neural-to-language mappings and the development of even more sophisticated models that can unlock the full potential of brain-computer interfaces. In doing so, we move closer to a future where the barriers between thought and expression are significantly diminished, enabling more direct and meaningful communication between minds and machines.

\section{Future work}

This paper is our first step of integrating diverse tasks within a single model, aiming to understand and generate neural signals across various modalities. For the next NeuGPT release, we plan to include ECoG and fMRI data to leverage their higher signal quality. We aim to refine the generation of continuous content such as speech and to introduce image modality for enhanced visualization. We also intend to explore new tasks, including diagnosis and intention detection, as shown in Fig. \ref{fig:future}.

\newpage

\section{Limitations}

Our study, innovative in its multi-modal language generation approach, is constrained by limited computational resources, which precluded the use of more extensive datasets.

\bibliography{anthology,custom}
\bibliographystyle{acl_natbib}


\appendix

\section{Appendix}
\label{sec:appendix}

\begin{figure*}[t]
    \centering
    \includegraphics[width=1\linewidth]{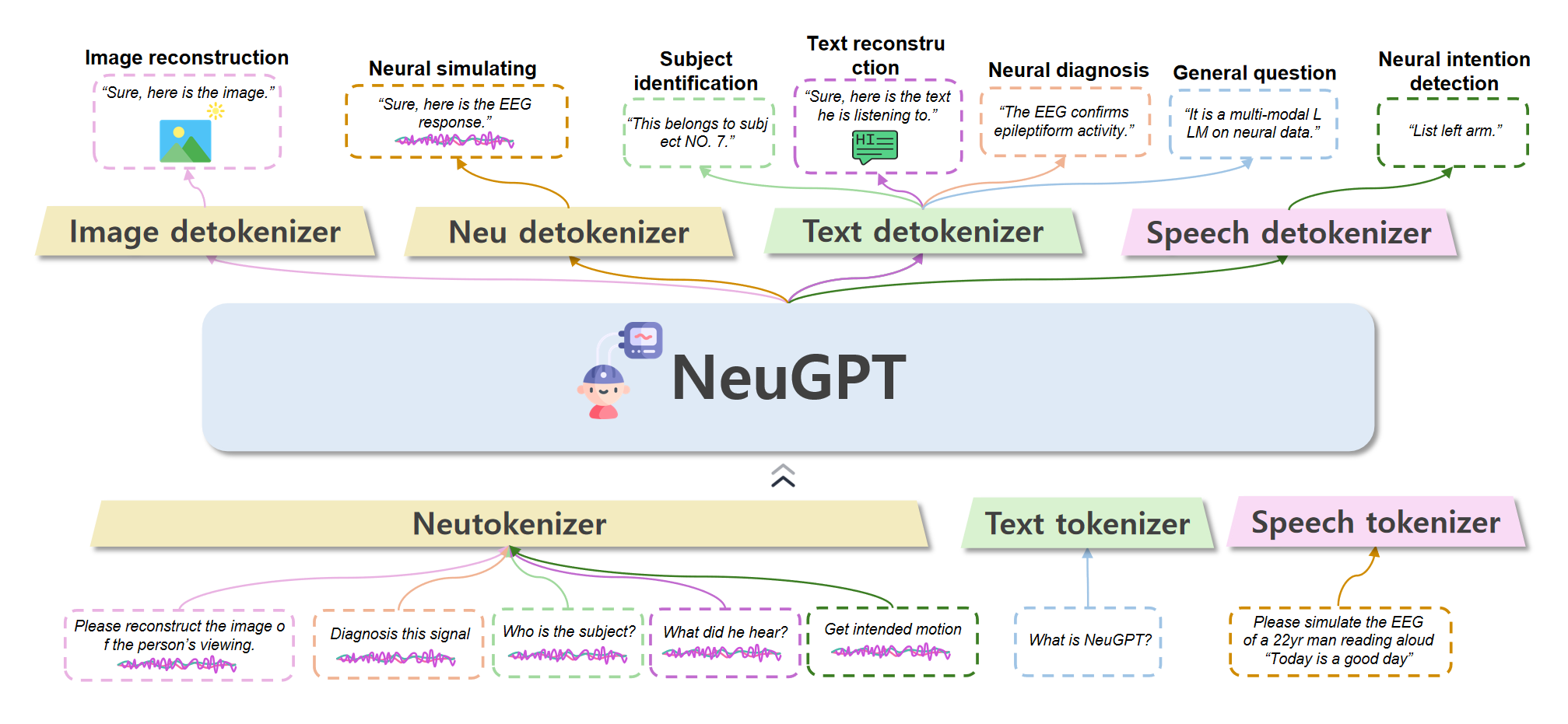}
    \caption{Future NeuGPT}
    \label{fig:future}
\end{figure*}

\end{document}